# INFLUENCE OF THE EVENT RATE ON DISCRIMINATION ABILITIES OF BANKRUPTCY PREDICTION MODELS

Lili Zhang[1], Jennifer Priestley[2] and Xuelei Ni[3]

[1]Program in Analytics and Data Science, Kennesaw State University, Georgia, USA
[2]Analytics and Data Science Institute, Kennesaw State University, Georgia, USA
[3]Department of Statistics, Kennesaw State University, Georgia, USA

*ABSTRACT*

*In bankruptcy prediction, the proportion of events is very low, which is often oversampled to eliminate this bias. In this paper, we study the influence of the event rate on discrimination abilities of bankruptcy prediction models. First the statistical association and significance of public records and firmographics indicators with the bankruptcy were explored. Then the event rate was oversampled from 0.12% to 10%, 20%, 30%, 40%, and 50%, respectively. Seven models were developed, including Logistic Regression, Decision Tree, Random Forest, Gradient Boosting, Support Vector Machine, Bayesian Network, and Neural Network. Under different event rates, models were comprehensively evaluated and compared based on Kolmogorov-Smirnov Statistic, accuracy, F1 score, Type I error, Type II error, and ROC curve on the hold-out dataset with their best probability cut-offs. Results show that Bayesian Network is the most insensitive to the event rate, while Support Vector Machine is the most sensitive.*

## 1. INTRODUCTION

The bankruptcy prediction has been studied for decades to support business operations and strategies with reliable counterparties [3]. For example, banks use the bankruptcy prediction as a part of their decision-making system to approve loans to corporates that are less likely to default. Investors foresee the bankruptcy possibility of organizations before making investments to ensure that they can get the most revenues.

To improve the discrimination ability of bankruptcy prediction and better differentiate bankruptcy instances and non-bankruptcy instances, researchers and practitioners have pursued two primary paths of study. First, explore significant variables for bankruptcy prediction. For example, financial ratio variables have been comprehensively studied and shown their predictive abilities. Second, enhance existing statistical models and machine learning algorithms or build novel ones for classifying bankruptcies and non-bankruptcies, by benefiting from the development of both algorithm theories and computation infrastructure. Moreover, considering that frequently the proportion of bankruptcy cases is substantively lower than the proportion of non-bankruptcies, the appropriate sampling for increasing the proportion of events (i.e. bankruptcies) also helps eliminate the imbalance bias and improve the performance of bankruptcy prediction models.

In this paper, we make contributions from all those three perspectives, namely, significant predictors, models, and sampling. The impacts of public records and firmographics variables on the bankruptcy prediction were explored to add values to widely used financial ratio variables. Both univariate analysis and multivariate analysis were conducted to measure their statistical association and significance. With significant variables as input variables, seven classification models were developed, including Logistic Regression, Decision Tree, Random Forest, Gradient Boosting, Support Vector Machine, Neural Network, and Bayesian Network, under different event rates. The event rate was oversampled from 0.12% to 10%, 20%, 30%, 40%, and 50% respectively, while the non-event rate was undersampled from 99.88% to 90%, 80%, 70%, 60%, and 50% simultaneously. Kolmogorov-Smirnov Statistic (i.e. K-S statistic) was used as the discrimination measure on how strong a model differentiates between events and non-events, under different event rates. Models were further evaluated and compared based on the overall classification accuracy, F1 score, Type I error, Type II error, and ROC curve, with their best probability cut-offs. All performance measures of the models were computed on the hold-out dataset.

The paper is organized into 7 sections. In Section 2, related work is reviewed. In Section 3, the data preprocessing is described. In Section 4, the statistical association and significance between predictors and the dependent variable is examined. In Section 5, the event rate is oversampled, and models are developed, diagnosed, evaluated, and compared. In Section 6, conclusions are made. In Section 7, future work is discussed.

## 2. RELATED WORK

Because of its importance in business decisions like investment and loan lending, the bankruptcy prediction problem has been studied through deriving significant predictors and developing novel prediction models. Altman proposed a set of traditional financial ratios, including Working Capital/Total Assets, Retained Earnings/Total Assets, Earnings before Interest and Taxes/Total Assets, Market Value Equity/Book Value of Total Debt, and Sales/Total Assets, and used them in the multiple discriminant analysis for the corporate bankruptcy prediction [2]. Those financial ratios were widely adopted and extended later [13] [4]. Amir came up with some novel financial ratio indicators, including Book Value/Total Assets, Cashflow/Total Assets, Price/Cashflow, Rate of Change of Stock Price, and Rate of Change of Cashflow per Share, in addition to Altman's ones, for a neural network model, and increased the prediction accuracy by 4.04% for a three-year-ahead forecast [4]. Everett et al. studied the impact of external risk factors (i.e. macro-economic factors) on small business bankruptcy prediction and proposed a logistic regression model [7]. Chava et al. demonstrated the statistical significance of industry effects by grouping firms into finance/insurance/real estate, transportation/communications/utilities, manufacturing/mineral, and miscellaneous industries [6].

From the methodology perspective, various statistical methods, machine learning algorithms, and hybrid models have been applied and compared for the bankruptcy prediction problem. Odom et al. proposed the first neural network model for bankruptcy prediction [13]. Zhang et al. showed that the neural network performed better than logistic regression and were robust to sampling variations [17]. Shin et al. found that the support vector machine outperformed the neural network on small training datasets [14]. Min et al. applied support vector machine with optimal kernel function hyperparameters [12]. Zibanezhad showed the acceptable prediction ability of decision tree on the bankruptcy prediction problem and determined the most important financial ratios [8]. Zikeba et al. proposed and evaluated a novel gradient boosting method for learning an ensemble of trees [18]. Sun et al. studied the application of Bayesian network on the bankruptcy prediction problem in respects of the influence of variable selection and variable discretization on the model performance [15]. Ahn et al. presented a hybrid methodology by combining rough set theory and neural network [1]. Huang et al. proposed a hybrid model by incorporating static and trend analysis in the neural network training [9]. Kumar et al. provided a comprehensive review on both the financial ratio variables and methods used for the bankruptcy prediction from 1968 to 2005, discussed merits and demerits of each method, and listed some important directions for future research [11]. Bellovary et al. reviewed 165 existing studies for the bankruptcy prediction and made some suggestions, where one suggestion was that the model accuracy was not guaranteed with the number of factors [5].

Most models proposed for bankruptcy prediction in the literature were directly developed on the dataset with a balanced proportion of bankruptcy and non-bankruptcy observations. However, data imbalance is a common issue in practice. Kim et al. proposed a geometric mean based boosting algorithm to address the data imbalance problem in the bankruptcy prediction, but only compared it with other boosting algorithms to show its advantage [19]. Zhou studied the effect of sampling methods for five bankruptcy prediction models, but the influence of event rates after resampling were not analyzed [20].

The models applied to the bankruptcy prediction utilize a variety of algorithms like Logistic Regression, Decision Tree, Random Forest, Gradient Boosting, Support Vector Machine, and Neural Network. The classification mechanisms behind these algorithms are different.

Logistic Regression formulates a function between the probability of the event ($\hat{p}$) and input variables ($x_1, x_2, \ldots, x_n$) defined as:

$$\hat{p} = \frac{1}{1 + e^{-(\beta_0 + \beta_1 x_1 + \cdots + \beta_n x_n)}}$$

The coefficients ($\beta_1, \beta_2, \ldots, \beta_n$) in the function are estimated by optimizing the maximum likelihood function defined as below, where $y$ is the actual value with the event denoted as 1 and the nonevent denoted as 0 [16].

$$\max y \log \hat{p} + (1 - y) \log(1 - \hat{p})$$

Decision Tree defines hierarchical rules by searching for optimal splits on input variables based on the Entropy or Gini index. The Entropy and Gini index of an input variable are defined below, where $x$ is a given input variable, $1, \ldots, k$ are levels in the dependent variable, and $p(i|x)$ is the conditional probability for the dependent variable taking value $i$ given $x$ [16].

$$Entropy(x) = -\sum_{i=1}^{k} p(i|x) \log_2\bigl(p(i|x)\bigr)$$

$$Gini(x) = 1 - \sum_{i=1}^{k} [p(i|x)]^2$$

Random Forest and Gradient Boosting are an ensemble of multiple decision tree models through bagging and boosting, respectively. In Random Forest, each tree is trained independently on a bootstrap dataset created from the original training dataset and then combined to a single prediction model by taking the average of all trees. In Gradient Boosting, each tree is trained sequentially based on a modified version of the original training dataset by utilizing the information of previously trained trees [10]. In tree-based models, a summary of variable importance can be obtained. The importance of each input variable is measured based on the Entropy or Gini reduction by splitting a given input variable. The larger the value is, the more important an input variable is.

Support Vector Machine defines a hyperplane for two-class classification by maximizing the marginal distance [10]. To handle the nonlinear relationship, a kernel function can be first applied to project the input variables to a higher feature space. Neural Network learns the relationship between the dependent variable and input variables by first transforming input variables with an activation function (Tanh, Sigmoid, etc.) through each hidden unit in one or more hidden layers and then adjusting the weights through backpropagation iteratively to minimize a loss function [22]. Bayesian Network represents the probability relationship and conditional dependencies between the dependent variable and input variables via a directed acyclic graph [23].

## 3. DATA

The public records, firmographics, and bankruptcy information of 11,787,287 U.S. companies in the 4[th] Quarter of both 2012 and 2013 were collected from a national credit reporting agency, and approved for use in this study. In the data, each corporate is identified by unique Market Participant Identifier (i.e. MPID). Public records include judgments and liens reported. Firmographics include industry, location, size, and structure. The bankruptcy indicator indicates whether a corporate is in bankruptcy or in business at the capture time point.

From the data, we aim to answer the following question explicitly, which can provide decision makers with insights into improved bankruptcy prediction.

*Given the public records and firmographics indicators of an organization, can we predict its operation status one year ahead?*

To address the question above, the dependent variable Bankruptcy Indicator Change (i.e. BrtIndChg) was created, as shown in Table 1. The original Bankruptcy Indicator (i.e. BrtInd) has two levels, 0 and 1, where 0 indicates that the organization is in operation and 1 indicates the bankruptcy. If an organization in business in 2012 went to bankruptcy in 2013, then BrtIndChg was assigned to 1. If the organization was still in business in 2013, then BrtIndChg was assigned to 0.

The raw data was cleaned and transformed prior to modeling, to address missing values, abnormal/incorrect values, and correlated variables. The following steps were applied to the data.

(1) Only keep observations with the level value 0 in the original 2012 BrtInd.
(2) Create the dependent variable BrtIndChg by comparing BrtInd in the dataset of 2012 and 2013 as shown in Table 1.
(3) Drop interval variables if the percentage of coded values or missing values is greater than 30%. A value of 30% was selected to optimize the percent of variance explained in the dataset.
(4) Drop observations in an interval variable or a categorical variable if the percentage of the abnormal/incorrect values in that variable is less than 5%.
(5) Discretize continuous variables into nominal variables. For example, the variable Number of Current Liens or Judgment was binned into Current Liens or Judgment Indicator (i.e. curLiensJudInd) with two levels, 0 and 1, where 0 means an organization does not have a lien or judgment currently and 1 means an organization has one or more liens or judgments currently.
(6) Retain the variable with the best predictive ability among several correlated variables. For example, based on both the variable definition and the Chi-Square value, the following variables are correlated: Current Liens/Judgment Indicator, Number of Current Liens/Judgment and Total Current Dollar Amounts on All Liens/Judgments. After comparing their performance, only the variable Current Liens/Judgment Indicator was kept.

Table 1. Creation of the Dependent Variable BrtIndChg

| BrtInd 2012 | BrtInd 2013 | BrtIndChg |
|---|---|---|
| 0 | 1 | 1 |
| 0 | 0 | 0 |

After the data preprocessing, the variables in Table 2 were prepared ready for further analysis and modeling. As described above, the bankruptcy is a rare event, which can be further confirmed by the distribution of the dependent variable BrtIndChg, as shown in Table 3. In our dataset, there are 0.12% of observations going into bankruptcy from 2012 to 2013 and 99.88% of observations staying in business from 2012 to 2013. Because the proportion of event cases is much less than the proportion of nonevent cases, we need to consider oversampling the event rate to have sufficient event cases to train the model and achieve better performance, which will be discussed in detail in Section 5.

Table 2. Variables for Analysis and Modelling.

| Variable | Type | Description |
|---|---|---|
| MPID | Nominal | Market Participant Identifier |
| BrtIndChg | Binary | Bankruptcy Indicator Change |
| curLiensJudInd | Nominal | Current Liens/Judgment Indicator |
| histLiensJudInd | Nominal | Historical Liens/Judgment Indicator |
| Industry | Nominal | Industry |
| LargeBusinessInd | Nominal | Large Business Indicator |
| Region | Nominal | Geographical Region |
| PublicCompanyFlag | Nominal | Public Company Flag |
| SubsidiaryInd | Nominal | Subsidiary Indicator |
| MonLstRptDatePlcRec | Interval | Number of Months Since Last Report Date on Public Records |

Table 3. Frequency of Dependent Variable.

| BrtIndChg | Frequency | Percent (%) |
|---|---|---|
| 1 | 1031 | 0.12 |
| 0 | 843330 | 99.88 |

## 4. EXPLORATORY ANALYSIS

To examine the statistical association and significance between each individual input variable and the dependent variable, bivariate analysis was performed. The results of odds ratio and Chi-square test can be found in Table 4. Based on the Chi-Square results, all the variables are significantly associated with the dependent variable except the variable PublicCompanyFlag. Based on the odds ratio, we have the following observations regarding their relationship:

- Current Lien/Judgment Indicator: The organizations which currently do not have any lien/judgment is about 47.1% less likely to go into bankruptcy in the following year than those which currently have liens or judgments.
- Historical Lien/Judgment Indicator: The organizations which did not have any lien/judgment is about 32% less likely to go into bankruptcy in the following year than the ones which historically had liens or judgments.
- Large Business Indicator: The organizations which are not large are about 45.8% less likely to go into bankruptcy in the following year than the ones which are large.
- Subsidiary Indicator: The organizations which are not subsidiaries are 74.5% more likely to go into bankruptcy in the following year than those organizations which are subsidiaries.
- Industry: By using the industry group 8 as the reference level, the organizations in the industry group 3 is about 2 times more likely going to the bankruptcy in the following year than the ones in the industry group 8.
- Region: By using the region group 9 as the reference level, the organizations in the region group 2 are about 55.7% less likely to go into bankruptcy in the following year than the ones in the region group 9.
- Number of Months Since Last Report Date on Public Records (i.e. MonLstDatePlcRec): Figure 1 shows that the distribution of MonLstDatePlcRec is very different in different levels of BrtIndChg, indicating their strong relationship.

Table 4. Univariate Odds Ratio and Chi-Square p-value.

| Effect | Odds Ratio | 95% Confidence Interval | Chi-Square p-value |
|---|---|---|---|
| curLiensJudInd 0 vs 1 | 0.529 | [0.447, 0.627] | <.0001 |
| histLiensJudInd 0 vs 1 | 0.680 | [0.601, 0.768] | <.0001 |
| LargeBusinessInd N vs Y | 0.542 | [0.474, 0.620] | <.0001 |
| LargeBusinessInd U vs Y | 0.202 | [0.165, 0.249] | |
| PublicCompanyFlag N vs Y | 0.295 | [0.104, 0.838] | 0.065 |
| PublicCompanyFlag U vs Y | 0.370 | [0.138, 0.989] | |
| SubsidiaryInd N vs Y | 1.745 | [0.997, 3.053] | <.0001 |
| SubsidiaryInd U vs Y | 0.411 | [0.261, 0.648] | |
| Industry 1 vs 8 | 1.538 | [0.947, 2.496] | <.0001 |
| Industry 2 vs 8 | 3.085 | [1.118, 8.514] | |
| Industry 3 vs 8 | 2.079 | [1.545, 2.797] | |
| Industry 4 vs 8 | 1.971 | [1.365, 2.847] | |
| Industry 5 vs 8 | 1.648 | [1.136, 2.392] | |
| Industry 6 vs 8 | 2.421 | [1.704, 3.439] | |
| Industry 7 vs 8 | 1.386 | [1.033, 1.859] | |
| Industry 9 vs 8 | 1.348 | [1.012, 1.795] | |
| Industry 10 vs 8 | 0.885 | [0.216, 3.629] | |
| Industry U vs 8 | 0.473 | [0.343, 0.651] | |
| Region 1 vs 9 | 0.699 | [0.479, 1.019] | <.0001 |
| Region 2 vs 9 | 0.443 | [0.358, 0.549] | |
| Region 3 vs 9 | 0.627 | [0.505, 0.779] | |
| Region 4 vs 9 | 0.913 | [0.686, 1.215] | |
| Region 5 vs 9 | 0.636 | [0.525, 0.772] | |
| Region 6 vs 9 | 1.203 | [0.928, 1.558] | |
| Region 7 vs 9 | 1.084 | [0.875, 1.343] | |
| Region 8 vs 9 | 1.194 | [0.920, 1.549] | |
| MonLstRptDatePlcRec | 0.971 | [0.969, 0.973] | <.0001 |

## 5. METHODOLOGY

To examine the influence of the event rate on discrimination abilities of bankruptcy prediction models, the proportion of events in the collected dataset was first oversampled from 0.12% to 10%, 20%, 30%, 40%, and 50%, respectively, with the proportion of non-events undersampled from 99.88% to 90%, 80%, 70%, 60%, and 50% correspondingly. Each resampled dataset was then split into training dataset and validation dataset, where the training dataset was used for training models and the validation dataset was used as the hold-out dataset for evaluating the performance of models. Seven classification models were developed, including Logistic Regression, Decision Tree, Random Forest, Gradient Boosting, Support Vector Machine, Bayesian Network, and Neural Network. K-S statistic was used to measure how strong the models were for differentiating events and non-events. Further models were evaluated and compared based on the overall accuracy, F1 score, Type I error, Type II error, and ROC curve.

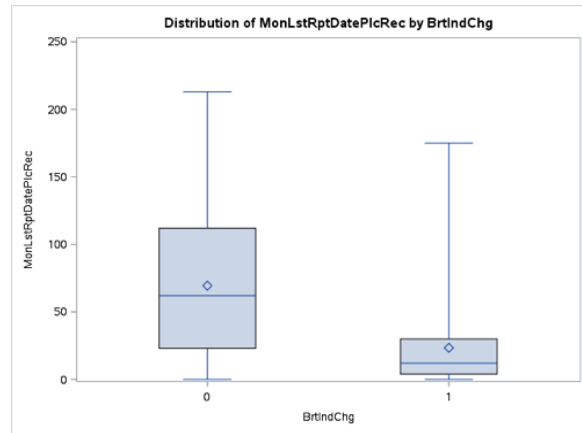

Figure 1. Boxplot of MonLstRptDatePlcRec by BrtIndChg

## 5.1. Sampling

The data sampling is done as follows.

(1) Event Rate Oversampling: The proportion of events in the dataset collected from the population is 0.12%, as indicated in Table 3. To avoid the model training biased towards non-events, the event rate in the data used for training and evaluating models should be increased. We keep all bankruptcy instances, and randomly select non-bankruptcy instances to adjust the proportions of events and non-events to 10% versus 90%, 20% versus 80%, 30% versus 70%, 40% versus 60%, and 50% versus 50%, respectively.

(2) Training Dataset and Validation Dataset Split: The out-of-sample test is used for evaluating models on the hold-out dataset. The originally collected dataset and resampled datasets are split into training and validation by 70% versus 30%, respectively.

## 5.2. Model Development and Evaluation

The models are developed using SAS Enterprise Miner 14.1. All variables in Table 4 are specified as initial inputs for all models. Every model is tuned to their best performance by searching different hypterparameter values. In Logistic Regression, backwards selection is used to select significant variables with the significance level set to 0.05. Decision Tree, Gradient Boosting, and Random Forest are all tree-based models. Entropy is used as the criteria of searching and evaluating candidate splitting rules for Decision Tree, while Gini index is used for Gradient Boosting and Random Forest. In Support Vector Machine, linear kernel function performs better than polynomial kernel function. In Neural Network, Tanh is used as the activation function in the hidden layer while Sigmoid is used in the output layer. There are 3 hidden units used in the hidden layer. In Bayesian Network, the significant variables are selected by G-Square with the significance level 0.2.

Table 5 summarizes K-S statistic of each model under different event rates, where K-S probability cut-offs are reported in the parenthesis. The larger K-S statistic is, the better a model differentiates between events and non-events. We have the following observations:

- When the proportion of events is 0.12%, Decision Tree, Gradient Descent, and Support Vector Machine have no discrimination ability at all, which means they classify all instances to non-bankruptcy. And the discrimination abilities of models Random Forest, Neural Network, and Logistic Regression are very small. However, Bayesian Network keeps good ability of differentiating between events and non-events.

- When the proportion of events is increased to 10%, Support Vector Machine still doesn't differentiate between events and non-events, while Decision Tree and Gradient Descent gain the discrimination abilities. Except Support Vector Machine, all the other models have K-S statistic around 0.5.

- When the proportion of events is increased to 20%, Support Vector Machine starts to have discrimination ability, but very small.
- When the proportion of events is increased to 30%, Support Vector Machine has similar K-S statistic as other models.
- Overall, the event rate influences discrimination abilities of models. For Support Vector Machine, as the event rate increases, its discrimination ability becomes better. For other models, they have slightly larger K-S statistic when the event rate is 10% and 50%.

Table 5. K-S Statistic (K-S Probability Cut-off) under Different Event Rates.

| Model \ Event (%) | 0.12 | 10 | 20 | 30 | 40 | 50 |
|---|---|---|---|---|---|---|
| Decision Tree | 0 (.) | 0.49 (0.06) | 0.424 (0.12) | 0.435 (0.21) | 0.475 (0.41) | 0.497 (0.43) |
| Gradient Boosting | 0 (.) | 0.495 (0.10) | 0.486 (0.25) | 0.471 (0.35) | 0.473 (0.45) | 0.532 (0.58) |
| Bayesian Network | 0.43 (0.12) | 0.501 (0.22) | 0.496 (0.33) | 0.473 (0.4) | 0.471 (0.44) | 0.503 (0.6) |
| Random Forest | 0.027 (0.02) | 0.536 (0.1) | 0.488 (0.25) | 0.46 (0.25) | 0.469 (0.43) | 0.516 (0.49) |
| Neural Network | 0.048 (0.01) | 0.523 (0.1) | 0.503 (0.2) | 0.488 (0.26) | 0.477 (0.51) | 0.519 (0.46) |
| Support Vector Machine | 0 (.) | 0 (.) | 0.02 (0.01) | 0.475 (0.43) | 0.439 (0.52) | 0.516 (0.54) |
| Logistic Regression | 0.037 (0.01) | 0.526 (0.08) | 0.502 (0.25) | 0.474 (0.25) | 0.44 (0.44) | 0.526 (0.56) |

Based on Table 5, when the event rate is very low, Support Vector Machine is the most sensitive and does not have the discrimination ability, while Bayesian Network is the most insensitive one and keeps moderate discrimination ability.

The models are further evaluated and compared based on overall accuracy, F1 Score, Type I error, and Type II error with their best probability cut-off, under the event rate 0.12% and 50%. The results are reported in Table 6 and 7. As shown in Table 6, except Bayesian Network, all the other models have very high Type II error. And for Bayesian Network, Type I error and Type II error seem to be okay, but F1 score is very small. If we further check its recall and precision, which are 82.69% and 0.28%, respectively. The low precision value indicates that the proportion of true bankruptcy instances is very small in the instances predicted to the bankruptcy.

Table 7 reports the performance measures of models in the scenario that we want to restrict Type II error of all models as close to 15% as possible for the comparison purpose, considering models give different performance measures with different probability cut-offs. For the bankruptcy prediction, Type II error is considered as a very important measure, because it costs more for misclassifying bankruptcy instances to non-bankruptcies. For Bayesian Network, its F1 score is increased substantively. The ensemble models Random Forest and Gradient Boosting perform very similar, better than Decision Tree. They give slightly larger accuracy, F1 score, and Type I error. The performance of Support Vector Machine is also good overall. All performance measures of Neural Network and Logistic Regression are very close, where Logistic Regression may be preferred for its high interpretability.

Table 6. Performance of Models under Event Rate 0.12%.

| Model | Accuracy | F1 Score | Type I Error | Type II Error | Cut-off |
|---|---|---|---|---|---|
| Decision Tree | 99.88% | . | 0% | 100% | . |
| Gradient Boosting | 99.88% | . | 0% | 100% | . |
| Bayesian Network | 64.43% | 0.0056 | 35.59% | 17.31% | 0.11 |
| Random Forest | 86.83% | 0.0022 | 13.08% | 87.95% | 0.01 |
| Neural Network | 99.23% | 0.0221 | 0.65% | 92.94% | 0.01 |
| Support Vector Machine | 99.88% | . | 0.00% | 100% | . |
| Logistic Regression | 99.41% | 0.0204 | 0.47% | 95.01% | 0.01 |

Table 7. Performance of Models under Oversampled Event Rate 50%.

| Model | Accuracy | F1 Score | Type I Error | Type II Error | Cut-off |
|---|---|---|---|---|---|
| Decision Tree | 72.26% | 0.7507 | 38.97% | 16.50% | 0.28 |
| Gradient Boosting | 73.44% | 0.7623 | 38.28% | 14.84% | 0.42 |
| Bayesian Network | 70.53% | 0.7413 | 43.41% | 15.53% | 0.37 |
| Random Forest | 73.93% | 0.7656 | 37.31% | 14.84% | 0.42 |
| Neural Network | 72.75% | 0.7579 | 39.81% | 14.70% | 0.37 |
| Support Vector Machine | 73.23% | 0.7605 | 38.56% | 14.98% | 0.49 |
| Logistic Regression | 72.61% | 0.7575 | 40.36% | 14.42% | 0.42 |

The performance difference of models can be further checked through ROC curves on the validation dataset, as shown in Figure 2 and Figure 3. In Figure 2, except Bayesian Network, ROC curves of other models are very close to the diagonal line which is the random prediction. Figure 3 shows that ROC curves of all models deviate from the diagonal line well, as the event rate is oversampled from 0.12% to 50%. And there is no large gap among their ROC curves.

Besides the performance measures, there are some other factors we may consider when selecting a model, like the variable importance and the model interpretability. The important variables determined by Decision Tree, Gradient Boosting, and Random Forest include MonLstDatePlcRec, Region, Industry, curLiensJudInd, histLiensJudInd, and LargeBusinessInd. Their importance measure can be found in Table 8. Note that for Decision Tree and Gradient Boosting, the importance measure presented here is the total Entropy or Gini reduction, while for Random Forest, the importance measure is the marginal Gini reduction. Logistic Regression is known for their high interpretability. The multivariate odds ratio and Chi-Square p-value of the resulting Logistic Regression model can be found in Table 9. The significant variables include curLiensJudInd, histLiensJudInd, LargeBusinessInd, Region, and MonLstDatePlcRec. Their multivariate odds ratio is consistent with their univariate odds ratio. For example, univariate odds ratio shows that curLiensJudInd is negatively associated with the dependent variable, which is the same as indicated by the multivariate odds ratio of curLiensJudInd.

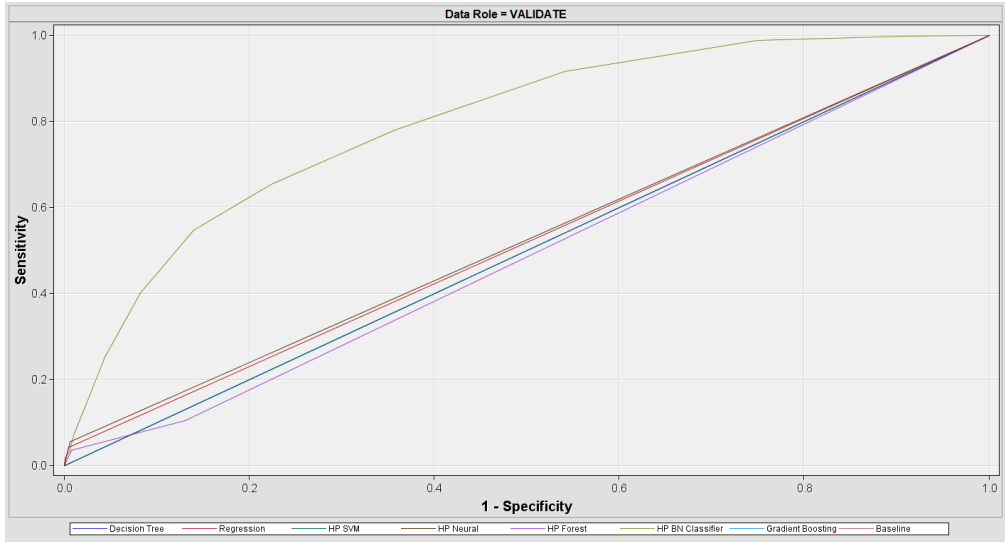

Figure 2. ROC Curve on Validation Dataset under Event Rate 0.12%

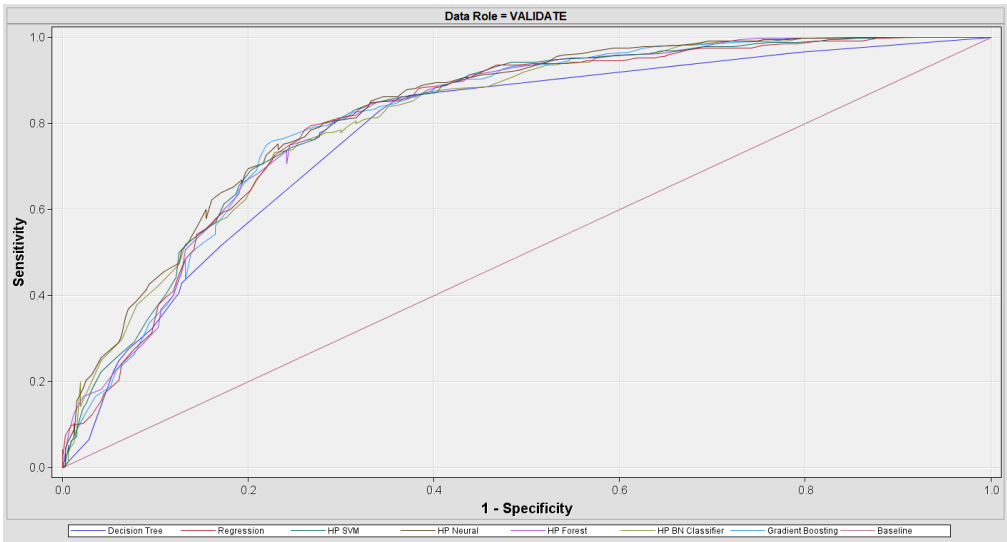

Figure 3. ROC Curve on Validation Dataset under Oversampled Event Rate 50%

Table 8. Variable Importance.

| Variable | Decision Tree | Gradient Boosting | Random Forest |
|---|---|---|---|
| MonLstRptDatePlcRec | 1.0000 | 1.0000 | 0.0911 |
| Region | 0.2423 | 0.2880 | 0.0048 |
| Industry | 0.1663 | 0.3516 | 0.0110 |
| curLiensJudInd | 0.1550 | 0.0820 | 0.0024 |
| histLiensJudInd | 0.1192 | 0.1205 | 0.0038 |
| LargeBusinessInd | 0.0308 | 0.2752 | 0.0100 |

Table 9. Multivariate Odds Ratio and Chi-Square p-value.

| Effect | Odds Ratio | Chi-Square p-value |
|---|---|---|
| curLiensJudInd 0 vs 1 | 0.573 | 0.0046 |
| histLiensJudInd 0 vs 1 | 0.508 | <.0001 |
| LargeBusinessInd N vs Y | 0.796 | <.0001 |
| LargeBusinessInd U vs Y | 0.332 | |
| Region 1 vs 9 | 1.067 | 0.0002 |
| Region 2 vs 9 | 0.411 | |
| Region 3 vs 9 | 0.583 | |
| Region 4 vs 9 | 0.839 | |
| Region 5 vs 9 | 0.558 | |
| Region 6 vs 9 | 0.858 | |
| Region 7 vs 9 | 0.881 | |
| Region 8 vs 9 | 1.261 | |
| MonLstRptDatePlcRec | 0.976 | <.0001 |

## 5.3. Probability Cut-off Tuning and Overfitting Checking

Classification models generate the predicted event probability, which ranges from 0 to 1, as the output. And probability cut-offs determine instances to be classified as events or non-events. With different probability cut-offs, the performance measures (accuracy, F1 score, Type I error, Type II error, etc.) of models will be different. They should be reported with their most appropriate probability cut-offs. Figure 4 shows some performance measures versus probability cut-offs of Logistic Regression under the event rate 50%. As shown, as the probability cut-off increases, the overcall classification rate (i.e. accuracy) increases first then decreases, the true positive rate (i.e. recall) decreases, the false positive rate (i.e. Type I error) decreases, and the true negative rate (i.e. specificity) increases. Because we want to keep Type II error as close to 15% as possible, which means the recall as close to 85% as possible, 0.42 is used as the probability cut-off, as highlighted by the vertical blue line.

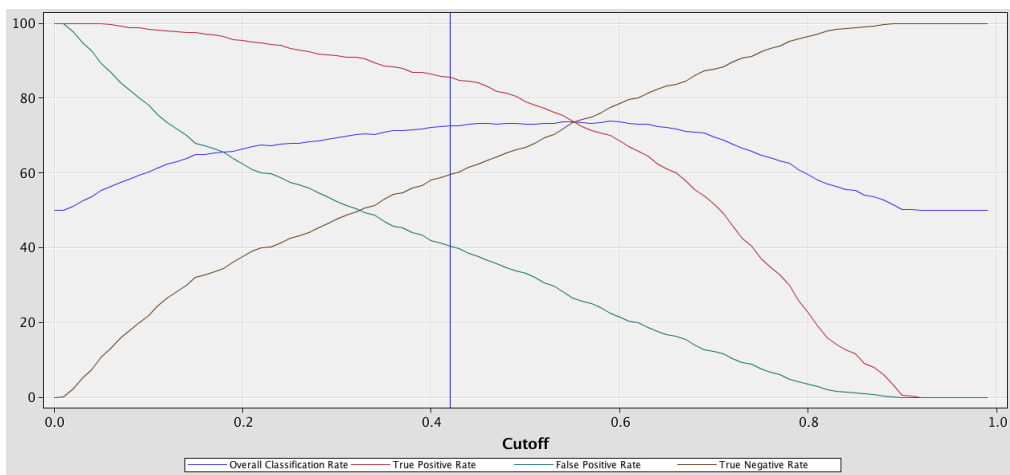

Figure 4. Performance Measures vs. Probability Cut-offs of Logistic Regression

Besides the probability cut-off, another issue we need to check with the model performance is the overfitting. Figure 5 shows ROC curves of models on the training dataset under oversampled event rate 50%. By comparing with Figure 3, we may conclude that there is no overfitting, because all models perform very similar on the training dataset and validation dataset.

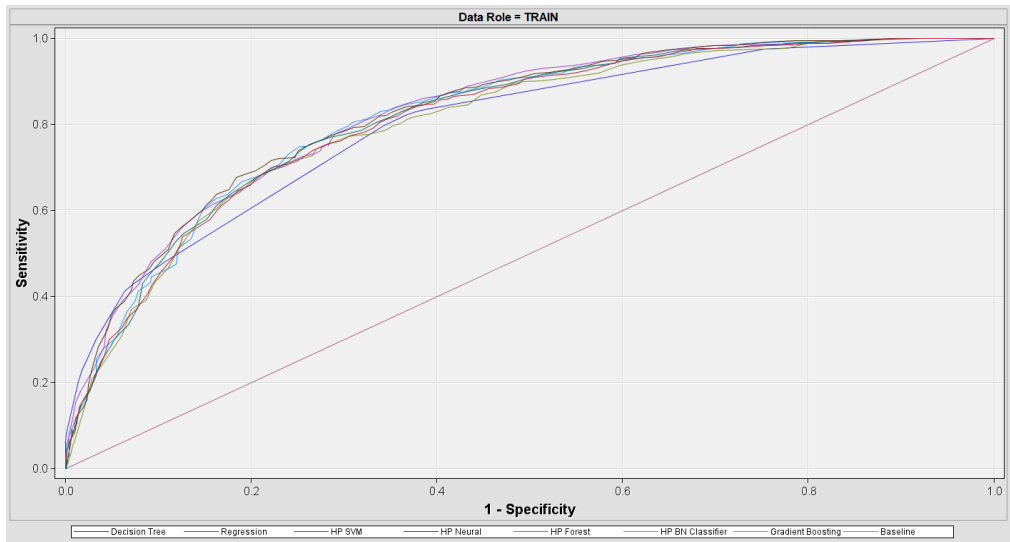

Figure 5. ROC Curve on Training Dataset under Oversampled Event Rate 50%

## 6. DISCUSSIONS AND CONCLUSIONS

Based on the univariate analysis and multivariate analysis, the impacts of public records and firmographics indicators were comprehensively studied. With them as input variables of different classification models, the model results show that public records and firmographics indicators play an important role in the bankruptcy prediction. This may serve as a reference for practitioners and researchers to include these information in the bankruptcy prediction model.

The event rate influences the performance of different classification models in different ways. When the event rate is very low, Support Vector Machine is the most sensitive one and does not have the discrimination ability, while Bayesian Network is the most insensitive one and keeps moderate discrimination ability. Support Vector Machine starts to differentiate events and non-events when the event rate is 20% and becomes much better as the event rate increases. Decision Tree and Gradient Boosting don't have the discrimination ability when the event rate is 0.12% but starts to gain the ability when the event rate is 10%. Except Support Vector Machine, all the other models have larger K-S statistic when the event rate is 10% and 50%.

Researchers and practitioners may examine the performance measures (K-S statistic, accuracy, F1 score, Type I error, Type II error, etc.) comprehensively and handle the tradeoffs among them as well as the model interpretability based on their expectations. If we only examine certain performance measures, the results may be misleading. For example, for Bayesian Network under the event rate 0.12%, its K-S statistic, Type I error and Type II error are good, but its accuracy, F1 score, and precision are not good, which means that lots of non-event instances are misclassified to be event instances. Another extreme example is that Support Vector Machine under the event rate 0.12% has the accuracy 100% but Type II error 100%, which indicates that all event instances are misclassified to be non-event instances. Moreover, different classification models generate quite different performance measures by using different probability cut-offs. The probability cut-off should be selected based on the scenario. In this paper, probability cut-offs are selected to make Type II error of models as close to 15% as possible for the comparison purpose.

Regarding the interpretability, Logistic Regression, Decision Tree and Bayesian Network might be favorable choices.

## 7. FUTURE WORK

In this study, we oversampled the event rate and undersampled the non-event rate by keeping all event instances and randomly selecting non-event instances to adjust their proportions. In the future, we may try different sample techniques like SMOTE [21] to balance the proportions of events and non-events and examine the influence further. Moreover, we only focused on the public records and firmographics indicators. Other information like financial ratios may be collected and included to improve the model performance as well as testing the model performance in a wider time span.

## Authors

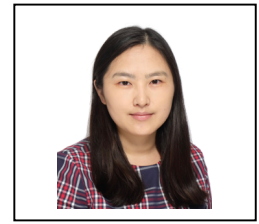

Lili Zhang is currently a Ph.D. candidate in Analytics and Data Science at Kennesaw State University, working on the data mining, machine learning, and data management.